\documentclass{article}
\usepackage{ijcai18}

\usepackage{times}
\usepackage{xcolor}
\usepackage{soul}
\usepackage[utf8]{inputenc}
\usepackage[small]{caption}

\usepackage{helvet}
\usepackage{courier}
\usepackage{amssymb}
\usepackage{verbatim}
\usepackage{amsmath, amsthm, amssymb}
\usepackage{graphicx,subfigure} 
\usepackage{multirow}
\usepackage{graphics,verbatim}
\usepackage{mathtools}
\usepackage{epstopdf,adjustbox}

\usepackage{tabularx}

\usepackage[hyperindex,breaklinks]{hyperref}       
\usepackage{url}            
\usepackage{booktabs}       
\usepackage{amsfonts}       
\usepackage{nicefrac}       
\usepackage{microtype}      

\usepackage{algpseudocode}
\usepackage{floatrow}
\usepackage{float}
\usepackage[flushleft]{threeparttable}

\usepackage{soul}

\title{Hashing over Predicted Future Frames for Informed Exploration of \\Deep Reinforcement Learning} %

\author{
Haiyan Yin,\; 
Jianda Chen,\; 
Sinno Jialin Pan 
\\
Nanyang Technological University, Singapore \\
\{haiyanyin, jianda001, sinnopan\}@ntu.edu.sg
}

\begin{document}

\maketitle

\begin{abstract}
In deep reinforcement learning (RL) tasks, an efficient exploration mechanism should be able to encourage an agent to take actions that lead to less frequent states which may yield higher accumulative future return. However, both \textit{knowing about the future} and \textit{evaluating the frequentness of states} are non-trivial tasks, especially for deep RL domains, where a state is represented by high-dimensional image frames. In this paper, we propose a novel informed exploration framework for deep RL, where we build the capability for an RL agent to predict over the future transitions and evaluate the frequentness for the predicted future frames in a meaningful manner.
To this end, we train a deep prediction model to predict \textit{future} frames given a state-action pair, and a convolutional autoencoder model to hash over the seen frames. In addition, to utilize the counts derived from the seen frames to evaluate the frequentness for the predicted frames, we tackle the challenge of matching the predicted future frames and their corresponding seen frames at the latent feature level.
In this way, we derive a reliable metric for evaluating the novelty of the future direction pointed by each action, and hence inform the agent to explore the least frequent one.
\end{abstract}

\section{Introduction}
Reinforcement learning (RL) involves an agent progressively interacting with an initially unknown environment, in order to learn an optimal policy with the objective of maximizing the cumulative rewards collected from the environment~\cite{sutton1998reinforcement}. Throughout the learning process, the RL agent alternates between two primal behaviors: \textit{exploration} - to try out novel states that could potentially lead to high future rewards; and \textit{exploitation} - to perform greedily according to the learned knowledge. In the past, exploitation of learned knowledge has been well studied, while how to efficiently explore through the state space is still remained as a critical challenge, especially for deep RL domains.

In deep RL domains, a state can be represented by low-level sensory inputs, such as image pixels~\cite{nature}, which is often high-dimensional or/and continuous. Thus, the state space for deep RL is huge and often intractable for searching. When performing exploration through such a huge state space, most existing single-task (e.g.,~\cite{Dueling,Priority,bellemare2016unifying}) and multi-task (e.g.,~\cite{Actor,Distillation,yin2017knowledge}) deep RL approaches adopt a simple exploration heuristic, $\epsilon$-greedy strategy, where the RL agent takes a random action with a probability of $\epsilon$, e.g., via uniform sampling for discrete-action domains~\cite{sutton1998reinforcement,nature} or corrupting action with i.i.d. Gaussian noise for continuous-action domains~\cite{lillicrap2015continuous}. In such a way, the agent explores the state space undirectedly~\cite{thrun1992efficient}, i.e., without incorporating any meaningful knowledge about the environment. Such exploration heuristic turns out to work well in simple problem domains but fails to handle more challenging domains, such as those with extremely sparse rewards that result in exponentially increasing state space.

Unlike the undirected exploratory behavior for agents using $\epsilon$-greedy strategy, when human beings are intending to explore an unfamiliar task domain, one often actively applies domain knowledge for the task, accounts for the state space that has been less frequently visited, and intentionally tries out actions that lead to novel states. In this work, we aim to mimic such exploratory behaviors to improve upon the $\epsilon$-greedy strategy with random action selection, and come up with a more efficient {\em informed} exploration framework for deep RL agents. On the one hand, we develop agent's knowledge on the environment and make it able to predict the future trajectories. On the other hand, we integrate the developed knowledge with hashing techniques over the high-dimensional state space in order to make the agent be able to realistically evaluate the novelty for the \textit{predicted} future trajectories.

Specifically, in our proposed informed exploration framework, first, we train an action-conditional prediction model to predict future frames given a state-action pair. Second, to perform hashing over the high-dimensional state space seen by the agent, we train a deep convolutional autoencoder to generate high-level features for the state and apply locality-sensitive hashing (LSH)~\cite{simHash} on the high-level state features to generate binary codes to represent each state. However, the learned hashing function is counting over the actually \textit{seen} states, while we need to query the counts for the \textit{predicted} future frames to compute their novelty. Hence, we introduce an additional training phase for the autoencoder to match the hash codes for the \textit{predicted} frames with that of their corresponding ground-truth frames (i.e., the actually \textit{seen} frames). In this way, we are able to utilize the environment knowledge and hashing techniques over the high-dimensional states to generate a reliable novelty evaluation metric for the future direction pointed by each action given a state.

\section{Related Work}
Recently, works on enhancing the exploration behavior of deep RL agents have demonstrated great potential in improving the performance of various challenging RL task domains. In~\cite{mnih2016asynchronous}, asynchronous training techniques are adopted and multiple agents are created to perform gradient-based learning to update the model parameters separately in the Atari 2600 domain. In~\cite{osband2016deep}, an ensemble of Q-functions is trained to reduce the bias for the values being approximated and increases the exploration depth in the Atari 2600 domain. In~\cite{houthooft2016vime}, an exploration strategy based on maximizing the information gain of the agent's belief about the environment dynamics is adopted for tasks with continuous states and actions. In~\cite{joshua2017ICLR}, the KL-divergence between the probability obtained from a learned dynamics model before and after updated by the experience is used to measure the surprise over the experience. In~\cite{bellemare2016unifying,Tang,martin2017count}, the novelty of a state is measured based on count-based mechanisms, and reward shaping is performed by adding a reward bonus term inferred from the value of counts. For all aforementioned approaches, the exploration strategy is incorporated either in the function approximation or during the optimization process, and the agent still needs to randomly choose action to explore without relying on any knowledge about the model. In this work, we aim to conduct informed exploration by utilizing the model-based knowledge to derive a deterministic choice of action for the agent to explore.

\paragraph{Exploration with Deep Prediction Models}
Recent works aiming to incentivize exploration via deep prediction models have shown promising results for deep RL domains. In~\cite{incentivize}, an autoencoder model is trained jointly with the policy model, and the reconstruction error from the autoencoder is used to determine the rareness of a state. In~\cite{pathak2017curiosity}, the prediction loss for the transited state at the feature level is used to infer the novelty of a state. In~\cite{ostrovski2017count}, a pixelCNN is trained jointly with the policy model as a density model for the state. The prediction gain of a state is measured as the difference of the state density given by the pixelCNN after and before observing that state. In the above approaches, the novelty of a state is measured by the loss or estimation output of another model, which is not exact statistics. In our work, we use the counts derived from hashing over the state space to reliably infer the novelty of a state. In~\cite{NIPS2015_5859}, which is a work mostly related to ours, an action-conditional prediction model is trained to predict the future frame given a state-action pair. Then they compute the Gaussian kernel distance between each predicted future frame and a set of history frames, and inform the agent to take the action that leads to the frames which are most dissimilar compared to a window of recently seen frames. In our work, we use the same frame prediction architecture as in~\cite{NIPS2015_5859}, but our work additionally performs hashing over the \textit{predicted} future frames to reliably infer the novelty of a possible action direction for the agent to explore.

\paragraph{Hashing for Deep RL Domain}
Running RL algorithms on the discretized features yields faster learning and promising performance. It is shown that the latent features learned by autoencoder trained in an unsupervised manner is of great promise to efficiently discretize the high-dimensional state space~\cite{blundell2016model}. In~\cite{Tang}, the state space for Atari 2600 domain is first discretized using the latent features derived from an autoencoder model. Then hashing is performed to encourage exploration by computing a reward bonus term in a form like MBIE-EB~\cite{strehl2008analysis}. Our work also introduces hashing over the state space based on latent features trained from deep autoencoder model, but the exploration mechanism is significantly different from~\cite{Tang}. First, in our work, we count over the actually \textit{seen} image frames, but query the hash for the \textit{predicted} frames. Second, in their approach, the reward bonus inferred from hashing is directly added to the function approximation target, which silently influences the previous states backwards through Bellman equation. But in our work, the count is used for informed exploration, which does not have direct influence on the approximated Q-values.

\section{Methodology}

\begin{figure*}[t!]
  \centering
  \includegraphics[width=0.84\columnwidth]{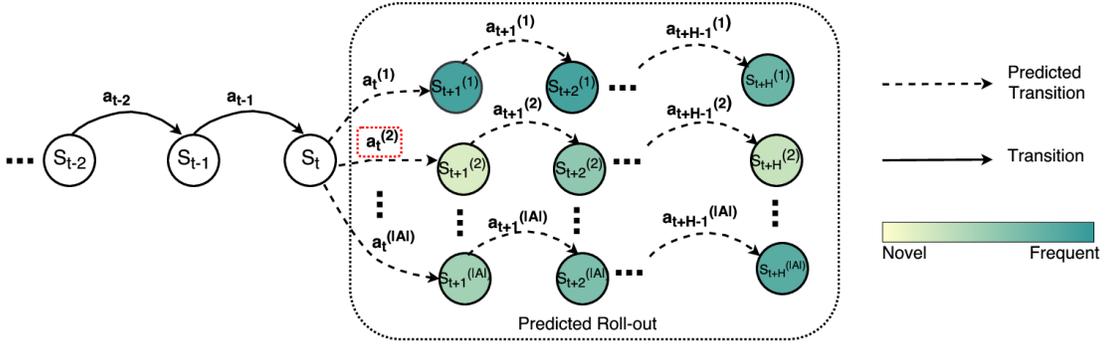} %
  \caption{An illustration over the decision making for the informed exploration framework. At state $S_t$, the agent needs to choose from $a_t^{(1)}$ to $a_t^{(|\mathcal{A}|)}$ an action to explore. Specifically, the states inside the dashed rectangle indicates \textit{predicted} future states. The color of circles after $S_t$ indicates frequency/novelty of states, where lighter ones indicate novel states and darker ones indicate frequent states. The agent first predicts future roll-outs, and then evaluate the novelty using hashing. In this example, $a_t^{(2)}$ is chosen because the followed roll-out is the most novel.
  }
  \label{fig:framework}
\end{figure*}
\subsection{Notations}
In this paper, we consider a discounted finite-horizon Markov Decision Process (MDP) with discrete actions. Formally, it is defined by a tuple $(\mathcal{S}, \mathcal{A}, \mathcal{P}, \mathcal{R}, \gamma)$, where $\mathcal{S}$ is a set of states which could be high-dimensional or continuous, $\mathcal{A}$ is a set of actions, $\mathcal{P}$ is a state transition probability distribution with $\mathcal{P}(s'|s,a)$ specifying the probability for transiting to state $s'$ after issuing action $a$ at state $s$, $\mathcal{R}$ is a reward function mapping each state-action pair to a reward in $\mathbb{R}$, and $\gamma\in[0,1]$ is a discount factor. The goal of the RL agent is to learn a policy $\pi$ that maximizes the expected cumulative future rewards given the policy: $\mathbb{E}_\pi[\sum_{t=0}^T\gamma^t\mathcal{R}(s_t,a_t)]$. In the context of deep RL, at each step $t$, an RL agent receives a state observation $\mathcal{S}_t\in\mathbb{R}^{r\times m\times n}$, where $r$ is the number of consequent frames to represent a state, and $m\times n$ is the dimension for each frame. The agent selects an action $a_t\in \mathcal{A}$ among all the $l$ possible choices, and receives a reward $r_t\in \mathbb{R}$.

\subsection{Informed Exploration Framework}
We propose an informed exploration framework to mimic the exploratory behavior of human beings under an unfamiliar task domain. The overall exploration decision making process is illustrated in Figure~\ref{fig:framework}.

Generally, the RL agent no longer randomly selects an action to explore without incorporating any domain knowledge. Instead, we aim to let the agent intentionally select the action that leads to the least frequent future states and thus explore the state space in an informed and deterministic manner. To this end, we build the capability of the RL agent on performing the following two tasks: 1) predicting over future transitions, and 2) evaluating the visiting frequency for those predicted \textit{future} frames.

\begin{figure}[h!]
\centering
\subfigure{\includegraphics[width=0.93\textwidth]{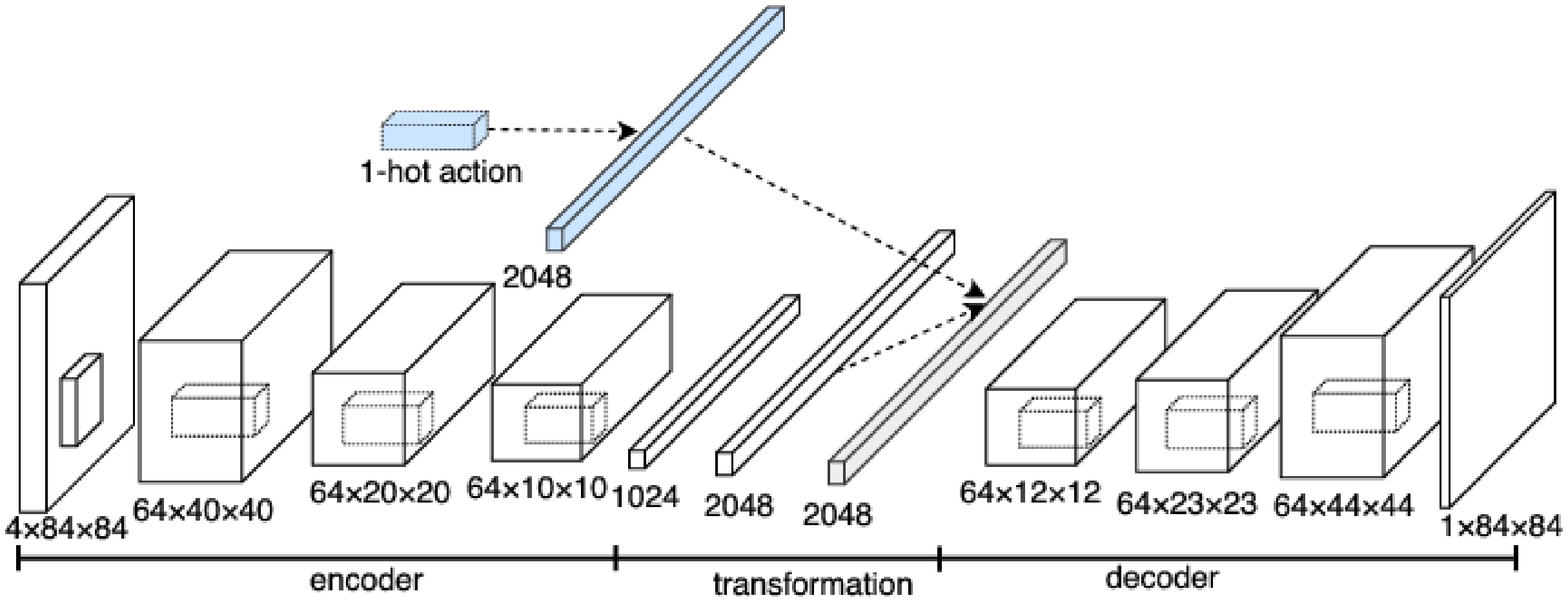}} \\

\subfigure{\hspace{-1.5mm}\includegraphics[width=0.83\textwidth]{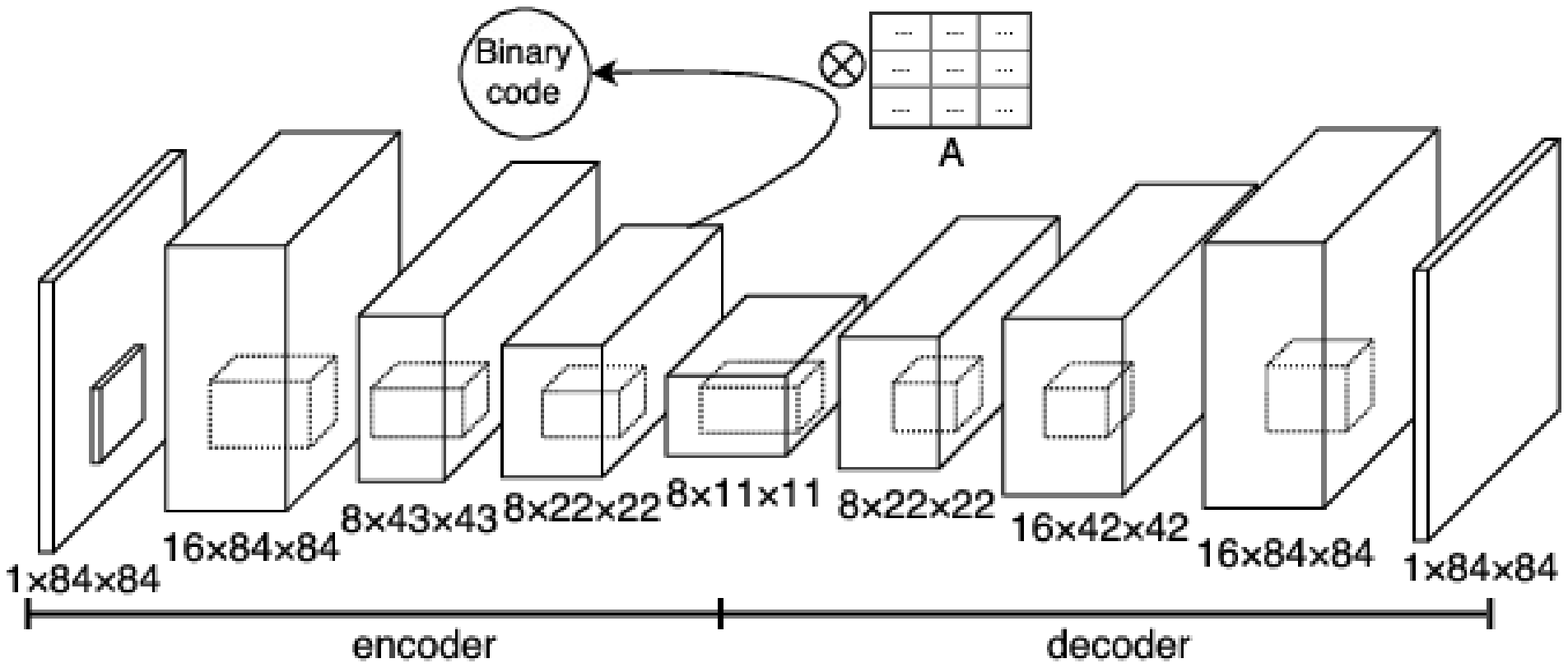}}  

\caption{Deep neural network architectures adopted for informed exploration. \textbf{Up:} action-conditional prediction model for predicting over future transition frames; \textbf{down:} autoencoder model for conducting hashing over the state space.}
\label{fig:predic}
\end{figure}

\subsubsection{Learning Transition Model with Prediction Network}
The architecture for action-conditional prediction is shown in Figure~\ref{fig:predic} (up). To be specific, we train a deep prediction model $f:(\mathcal{S}_t,a_t) \rightarrow \mathcal{S}_{t+1}$ to make the agent able to predict over the future transitions given a state-action pair. The state input $\mathcal{S}_t$ is a set of $r$ recent image frames, and the action input $a_t$ is represented by $\mathbf{a}_t \in \mathbb{R}^{l}$, which is a one-hot vector, where $l$ is the number of actions for the task domain. To predict a new state, the model predicts one single frame at a time, denoted as $\boldsymbol{\hat{s}}\in \mathbb{R}^{m\times n}$. The new state $\mathcal{S}_{t+1}$ is formed by concatenating the predicted new frame with the most recent $r\!-\!1$ frames.  We adopt the action-conditional transformation as proposed in~\cite{NIPS2015_5859} to form a joint feature for the state input and the action input. Specifically, the state input is first passed through three stacked convolutional layers to form a feature vector $\mathbf{h}_t^{s} \in \mathbb{R}^{h}$. Then the state feature $\mathbf{h}_t^s$ and the one-hot action feature $\mathbf{a}_t$ perform a linear transformation by multiplying with their corresponding transformation matrix $\mathbf{W}^s_t\in \mathbb{R}^{k\times h}$ and $\mathbf{W}^a_t\in \mathbb{R}^{k\times l}$. After the linear transformation, both features are shaped with the same dimensionality. Then the features for the state and action after the linear transformation performs a multiplicative interaction to form a joint feature as follows,
\begin{equation}
\mathbf{h}_t = \mathbf{W}_t^s\mathbf{h}_t^s \odot \mathbf{W}_t^a\mathbf{h}_t^a. \nonumber
\end{equation}
Afterwards, the joint feature $\mathbf{h}_t$ is passed through stacked deconvolutional layers and a sigmoid layer to form the final prediction output. To predict over multiple future steps, the prediction model progressively composes the new state using its prediction result to predict the next-step transition.
Note that our prediction on frame-to-frame level is more precise than those on feature level because the prediction target of frame-to-frame is exact image from game environment while feature prediction target is encoded feature extracted from a trained neural network.

\subsubsection{Hashing over the State Space with Autoencoder and LSH}
To evaluate the novelty of a state, we adopt a hashing model to count over the state space. We first train an autoencoder model on frames, $g\!:\! \mathbf{s} \!\in\! \mathbb{R}^{m\times n} \!\rightarrow\! \boldsymbol{\hat{s}} \!\in\! \mathbb{R}^{m\times n}$, in an unsupervised manner (to \textit{classify} the pixels), with the reconstruction loss defined as follows~\cite{kingma2013auto},
\begin{equation}\label{eq:loss:ae}
\mathcal{L}_{\mbox{rec}}(\mathbf{s}_t) = -\frac{1}{m n} \sum_{j=1}^{n}\sum_{i=1}^{m}\left( \mbox{log}\; p(\boldsymbol{\hat{s}}_{t_{ij}})\right),
\end{equation}
where $\boldsymbol{\hat{s}}_{t_{ij}}$ is the reconstructed pixel at the $i$-th row and the $j$-th column. The architecture for the autoencoder model is shown in Figure~\ref{fig:predic} (down). To be specific, each convolutional layer is followed by a Rectifier Linear Unit (ReLU) layer and a max pooling layer with a kernel size $2\times 2$. To discretize the state space, we hash over the last frame $s_t$ of each state. We adopt the output of the last ReLU layer from the encoder as the high-level state features, and denote by $\phi(\cdot)$ the corresponding feature map that generates the high-level feature vector $\mathbf{z}_t \!\in\! \mathbb{R}^d$ of a state, i.e., $\phi(\mathbf{s}_t)=\mathbf{z}_t$. To further discretize the state feature, locality-sensitive hashing (LSH)~\cite{simHash} is adopted upon $\mathbf{z}_t$. To this end, a projection matrix $\mathbf{A}\in\mathbb{R}^{p\times d}$ is randomly initialized with i.i.d. entries drawn from a standard Gaussian $\mathcal{N}(0,1)$. Then by projecting feature $\mathbf{z}$ through $\mathbf{A}$, the sign of the outputs form a binary code, $\mathbf{c} \in \mathbb{R}^p$. With the introduced discretization scheme, we are able to count over the state space for the problem domain. During the RL process, a hash table $\mathcal{H}$ is created. The count for a state $\mathcal{S}_t$, denoted by $\psi_t$, can be stored, queried and updated from the hash table. Overall, the process for counting over a state $\mathcal{S}_t$ is expressed in the following way:
\begin{equation}\label{Eq:count}
\mathbf{z}_t = \phi(\mathbf{s}_t), \;\; \mathbf{c}_t = \mbox{sgn}(\mathbf{A}\mathbf{z}_t), \; \mbox{ and } \; \psi_{t} = \mathcal{H}(\mathbf{c}_t).
\end{equation}

\subsubsection{Matching the Prediction with Reality}
To derive the novelty for the \textit{predicted} frames while updating the hash table by counting over the \textit{seen} frames, we need to match the predictions with realities, i.e., make the hash codes for the \textit{predicted} frames to be the same as their corresponding ground-truth \textit{seen} frames in training. To this end, we introduce an additional training phase for the autoencoder model $g(\cdot)$. To make the hash codes to be the same, the derived feature vectors of the predicted frames and the ground-truth seen fames through $\phi(\cdot)$ need to be close to each other. We introduce an additional loss function of a pair of ground-truth seen frame and predicted frame $(\mathbf{s}_t,\boldsymbol{\hat{s}}_t)$ as follows,
\begin{equation}\label{eq:loss:match}
\mathcal{L}_{\mbox{mat}}(\mathbf{s}_t,\boldsymbol{\hat{s}}_t) = \|\phi(\mathbf{s}_t) - \phi(\boldsymbol{\hat{s}}_t)\|_2 \\
\end{equation}
Finally, by combing (\ref{eq:loss:ae}) and (\ref{eq:loss:match}), we define the following overall loss function,
\begin{equation}\label{eq:loss:overall}
\mathcal{L}(\mathbf{s}_t, \hat{s}_t;\theta) = \mathcal{L}_{\mbox{rec}}(\mathbf{s}_t) + \mathcal{L}_{\mbox{rec}}(\boldsymbol{\hat{s}}_t) + \lambda\mathcal{L}_{\mbox{mat}}(\mathbf{s}_t, \boldsymbol{\hat{s}}_t),
\end{equation}
where $\theta$ is the parameter for the autoencoder. Note that even though the prediction model could generate almost identical frames, training the autoencoder with only the reconstruction loss may lead to distinct state codes in all the task domains (details will be shown in Section~\ref{Sec:Eval_hash}). Therefore, the effort for matching the codes is necessary.

However, matching the state code while guaranteeing a satisfying reconstruction behavior is extremely challenging. Fine tuning an autoencoder fully trained with $\mathcal{L}_{\mbox{rec}}$ by involving the additional code matching loss $\mathcal{L}_{\mbox{mat}}$ would fast disrupt the reconstruction behavior before the code loss could decrease to the expected level. Training the autoencoder from scratch with both $\mathcal{L}_{\mbox{rec}}$ and $\mathcal{L}_{\mbox{mat}}$ is also difficult, as $\mathcal{L}_{\mbox{mat}}$ is initially very low and $\mathcal{L}_{\mbox{rec}}$ is very high. The network can hardly find a direction to consistently decrease $\mathcal{L}_{\mbox{rec}}$ with such an imbalance. Therefore, in this work, we propose to train the autoencoder for two phases, where the first phase uses $\mathcal{L}_{\mbox{rec}}$ to train on \textit{seen} frames until convergence, and the second phase uses the composed loss function $\mathcal{L}$ as in (\ref{eq:loss:overall}) to address the requirement for matching the prediction with reality.

\subsubsection{Computing Novelty for States}
Once the prediction model $f(\cdot)$ and the autoencoder model $g(\cdot)$ are both trained, the agent could perform informed exploration as illustrated in Figure~\ref{fig:framework}. At each step, the agent performs exploration with a probability less than $\epsilon$ and performs greedy action selection otherwise. When performing exploration, the agent strategically selects the most novel action direction to explore. Given the state $\mathcal{S}_t$, the agent first performs roll-out with length $H$ for the future trajectories predicted by the prediction model, for all the possible actions $\mathbf{a}_j\in \mathcal{A}$. Formally, the novelty score for an action ${a}_j$ given state $\mathcal{S}_t$, denoted by $\rho({a}_j|\mathcal{S}_t)$, is computed as,
\begin{equation}
\rho({a}_j|\mathcal{S}_t) = \sum_{i=1}^{H}\frac{\beta^{i-1}}{\sqrt{\psi_{t+i}+0.01}},
\end{equation}
where $\psi_{t+i}$ is the count for the future state $\mathcal{S}_{t+i}$ derived from (\ref{Eq:count}) by hashing over the predicted frame $\hat{s}_{t+i}$, $H$ is a predefined roll-out length, and $\beta$ is a real-valued discount rate. After evaluating the novelty for all the possible actions, the agent selects the one with the highest novelty score to explore. Overall, the policy for the RL agent with the proposed informed exploration strategy is defined as:
\[
{a_t}=
    \begin{cases}
    \begin{split}
        & \arg\max\limits_{a} [\mathcal{Q}(\mathcal{S}_t, {a})] & p \ge \varepsilon, \\
        & \arg\max\limits_{a} [\rho({a}|\mathcal{S}_t)]         & p < \varepsilon, \\
    \end{split}
    \end{cases}
\]
where $p$ is a random value sampled from Uniform (0,1), and $\mathcal{Q}(\mathcal{S}_t,a)$ is the output of the Q-value function.

\section{Experiments}

In the empirical evaluation, we use the Arcade Learning Environment (ALE)~\cite{bellemare13arcade}, which consists of Atari 2600 video games as the testing domain. We choose 5 representative games that require significant exploration to learn the policy: \textit{Breakout}, \textit{Freeway}, \textit{Frostbite}, \textit{Ms-Pacman} and \textit{Q-bert}. Four of the five games, \textit{Freeway}, \textit{Frostbite}, \textit{Ms-Pacman} and \textit{Q-bert} are classified as \textit{hard} exploration games, based on the taxonomy of exploration proposed in~\cite{bellemare2016unifying}, where \textit{Freeway} has sparse reward and the others have dense reward. \textit{Breakout}, though not classified as hard exploration game, demonstrates significant exploration bottleneck with standard exploration attempt, as the state space is changing rapidly as agent learn, and the performance with standard exploration falls far behind advanced exploration technique. Hence it is also included as a test domain.

For all the tasks, we use the state representation that concatenates 4 consequent image frames of size $84\times 84$.

\begin{figure*}[t!]
\centering
\includegraphics[width=1.0\textwidth]{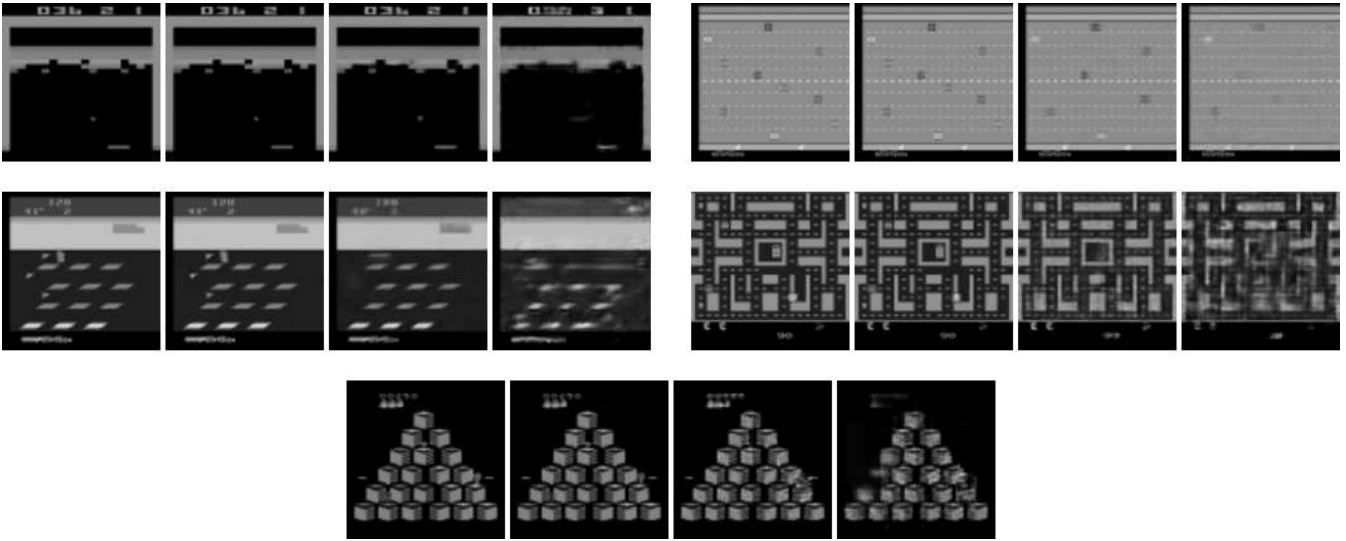}
 \caption{The prediction and reconstruction result for each task domain. For each task, we present 1 set of frames, where the four frames are organized as follows: (1) the \textbf{ground-truth} frame seen by the agent; (2) the \textbf{predicted} frame by the prediction model; (3) the \textbf{reconstruction} of autoencoder trained only with reconstruction loss; (4) the \textbf{reconstruction} of autoencoder trained after the second phase (i.e., trained with both reconstruction loss and code matching loss). Overall, the prediction model could perfectly produce frame output, while the fully trained autoencoder generates slightly blurred frames.
  }
    \label{fig:all_frames}

\end{figure*}

\subsection{Evaluation on Prediction Model}
The architecture of the prediction model is identical to the one shown in Figure~\ref{fig:predic} (up). To train the prediction model, we create a training dataset which consists of 500,000 transition records generated by a fully trained DQN agent performing under standard $\epsilon$-greedy, where $\epsilon$ is set equal to 0.3 (same as~\cite{NIPS2015_5859}). During training, we adopt Adam~\cite{adam} with a learning rate of $10^{-3}$ and a mini-batch size of 100. Moreover, we discount the gradient scale by multiplying the gradient value by $10^{-2}$.

We show the pixel prediction loss in mean square error (MSE) for multi-step future prediction in Table~\ref{table:pred:mse}. For all the task domains, the prediction losses are within a small scale. For multi-step prediction, as expected, the prediction loss increases with the increase of the prediction length. We demonstrate that the trained prediction models are able to generate realistic future frames which are visualized to be very close to the ground-truth frames in Figure~\ref{fig:all_frames}.

\begin{table}[ht!]
\footnotesize
  \caption{The prediction loss in MSE for the trained prediction model.}
  \label{table:pred:mse}
  \centering
  \scalebox{0.92}{
  \begin{tabular}{lllll}
    \toprule
    Game     & 1-step  &  3-step  & 5-step & 10-step \\
    \midrule
    \textbf{Breakout}    & 1.114e-05 &  3.611e-04 &     4.471e-04        &  5.296e-04  \\
    \textbf{Freeway}      & 2.856e-05 & 0.939e-05   & 1.424e-04     & 2.479e-04      \\
    \textbf{Frostbite}    & 7.230e-05 & 2.401e-04  & 5.142e-04  & 1.800e-03   \\
    \textbf{Ms-Pacman}   & 1.413e-04 &  4.353e-04  & 6.913e-04  & 1.226e-03   \\
    \textbf{Q-bert}   & 5.300e-05 & 1.570e-04  & 2.688e-04  & 4.552e-04    \\
    \bottomrule
  \end{tabular}
  }
\end{table}


\vspace{-3mm}
\subsection{Evaluation on Hashing with Autoencoder and LSH}
\label{Sec:Eval_hash}
The architecture for the autoencoder model is identical to that shown in Figure~\ref{fig:predic} (down). The autoencoder is trained on a dataset collected in an identical manner as that for the prediction model. It is trained under two phases. In the first phase, it is trained with only reconstruction loss. We use Adam as optimization algorithm, $10^{-3}$ as learning rate, a mini-batch size of 100 and discount the gradient by multiplying $10^{-2}$. In the second phase, we train the autoencoder based on the loss in (\ref{eq:loss:overall}). Specifically, we use Adam as optimization algorithm, $10^{-4}$ as learning rate, a mini-batch size of 100, and $\lambda$ value of 0.01. We discount the gradient by multiplying the gradient value by $5\times 10^{-3}$.

\begin{figure*}[t!]
\ffigbox[\textwidth][6.3cm]{%

\begin{subfloatrow}
  \ffigbox[\FBwidth][]
    {} 
    {\includegraphics[height=5.5cm]{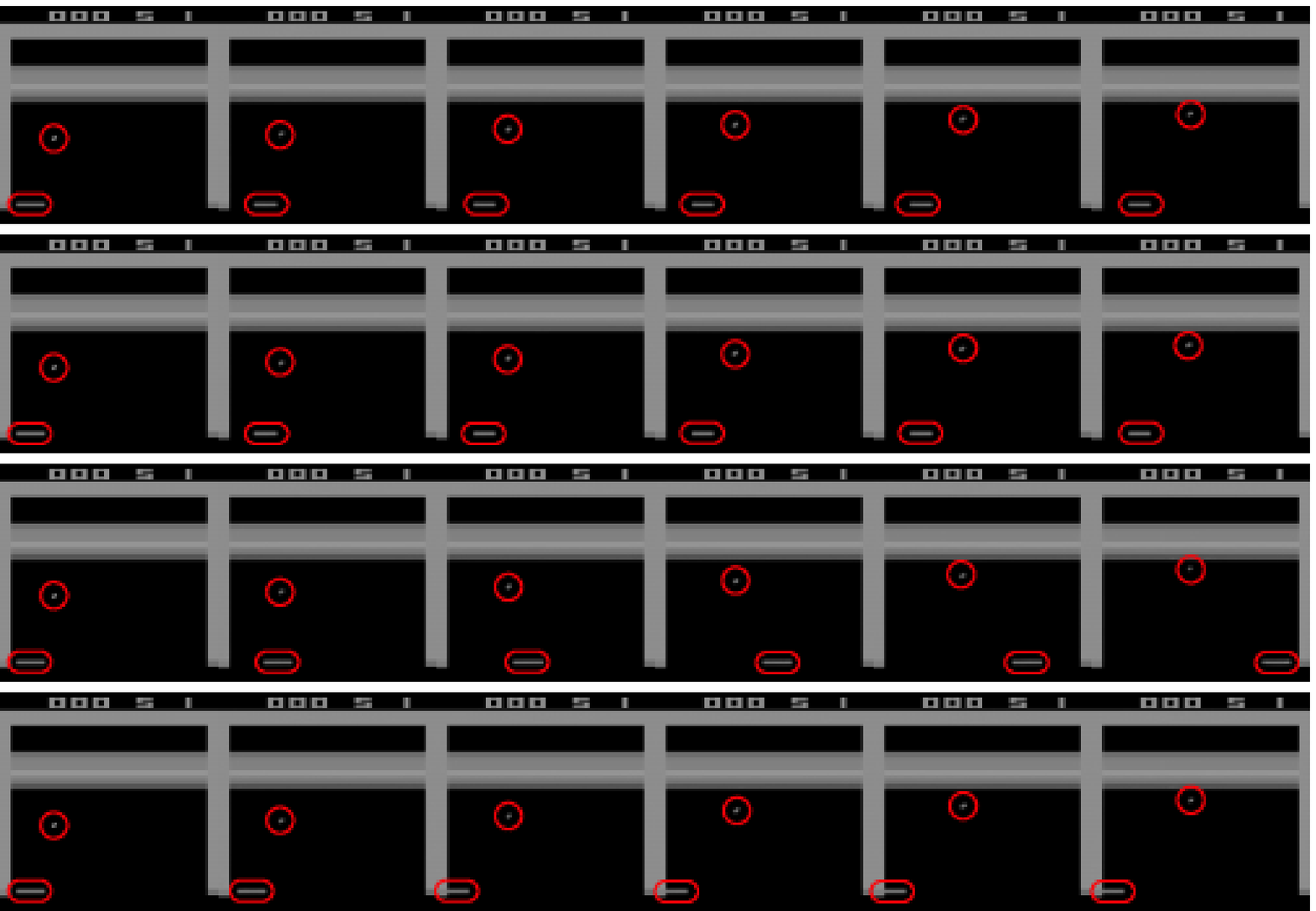}}

\begin{subfloatrow}
  \vbox to 5.5cm{%
    \ffigbox[\FBwidth][]{}
    {\includegraphics[width=8cm,height=1.3cm]{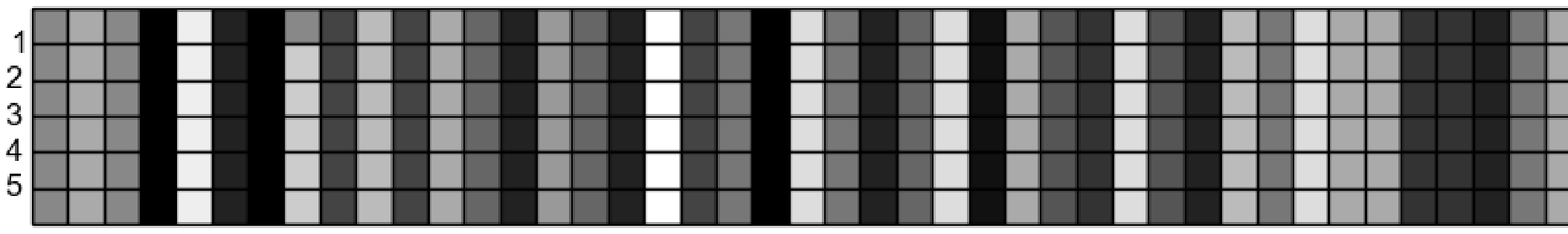}} \vss
    \ffigbox[\FBwidth][]{}
    {\includegraphics[width=8cm,height=1.3cm]{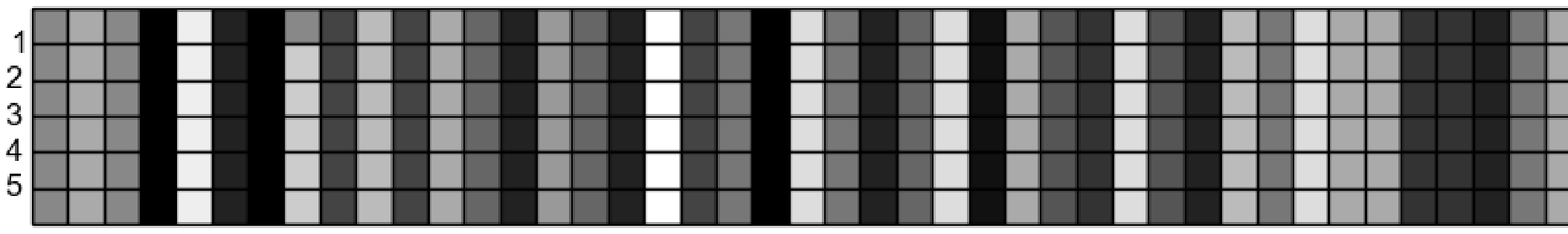}} \vss
    \vspace{4mm}
    \ffigbox[\FBwidth][]{}
    {\includegraphics[width=8cm,height=1.3cm]{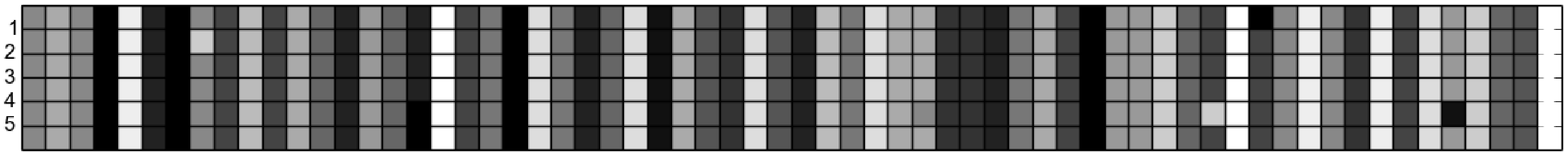}} \vss
    \ffigbox[\FBwidth][]{}
    {\includegraphics[width=8cm,height=1.3cm]{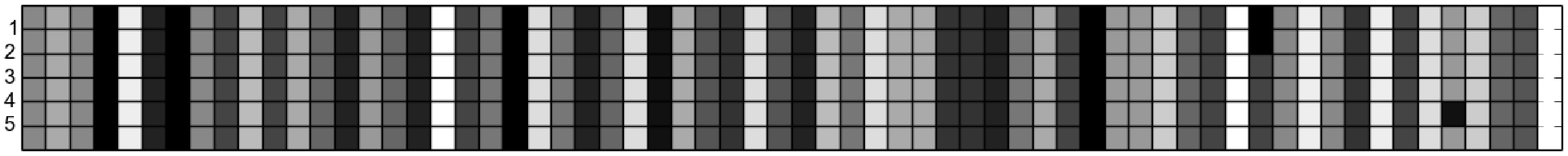}} \vss
  }
\end{subfloatrow}

\end{subfloatrow} 
}

 \captionsetup[subfigure]{justification=centering}
 \floatbox{figure}{%
  \vspace{0.5mm}
  \caption{%
  \textbf{Left}: the predicted future trajectories for each action in \textit{Breakout}. In each row, the first frame is the \textit{ground-truth} frame and the following five frames are the \textit{predicted} future trajectories with length 5. In each row, the agent takes one of the following actions (continuously): (1) no-op; (2) fire; (3) right; (4) left. \textbf{Right}: the hash codes for the frames in the same row ordered in a top-down manner. To save the space, four binary codes are grouped into one hex code, i.e., in a range of [0,15]. The color map is normalized linearly by hex value.
  }\label{fig:breakout:hash}
}

\end{figure*}

\begin{figure}[H]
\centering

\subfigure[Code Loss]{

  \includegraphics[width=0.48\textwidth]{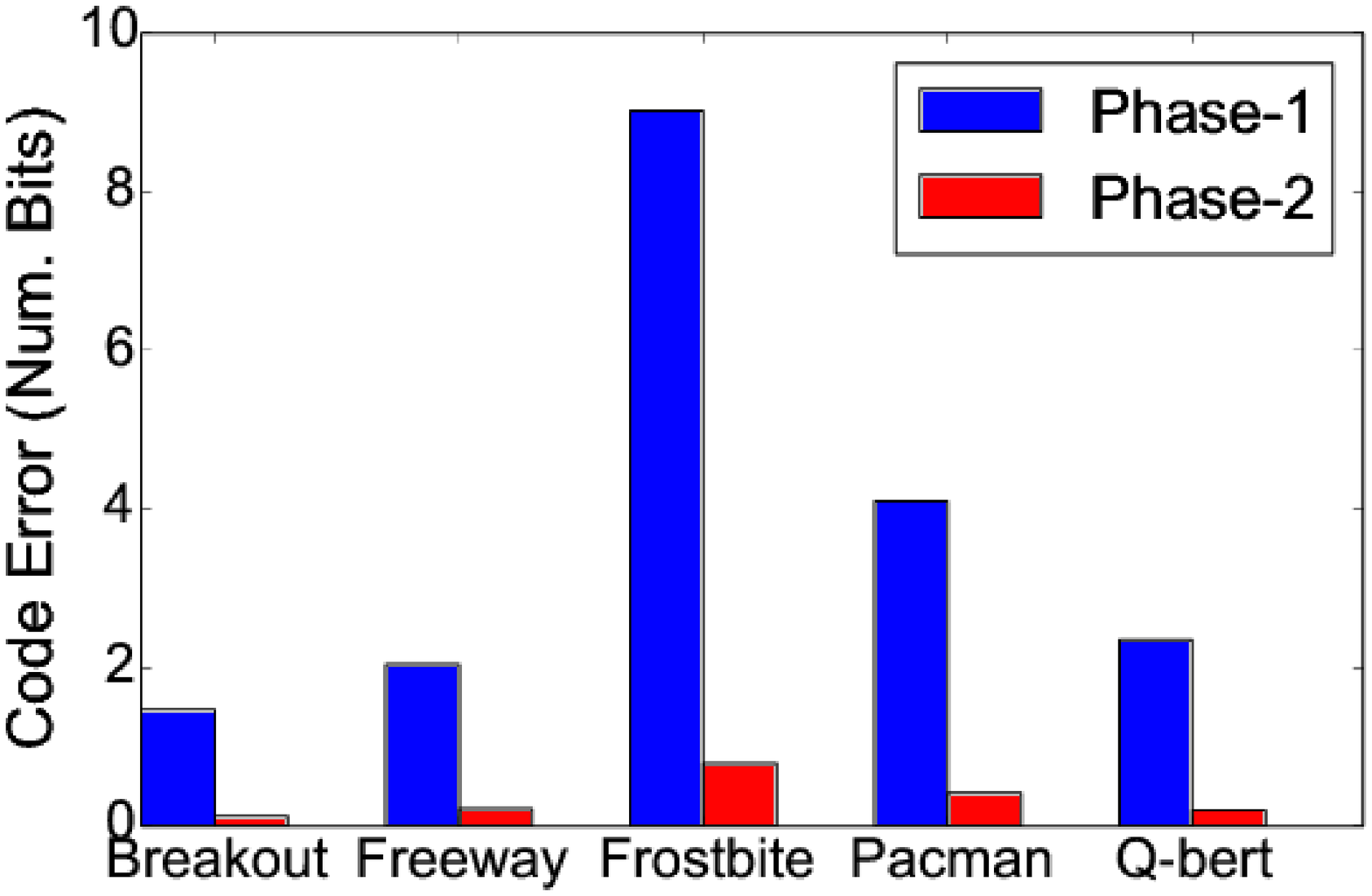}}
\subfigure[Reconstruction Loss]{

  \includegraphics[width=0.48\textwidth]{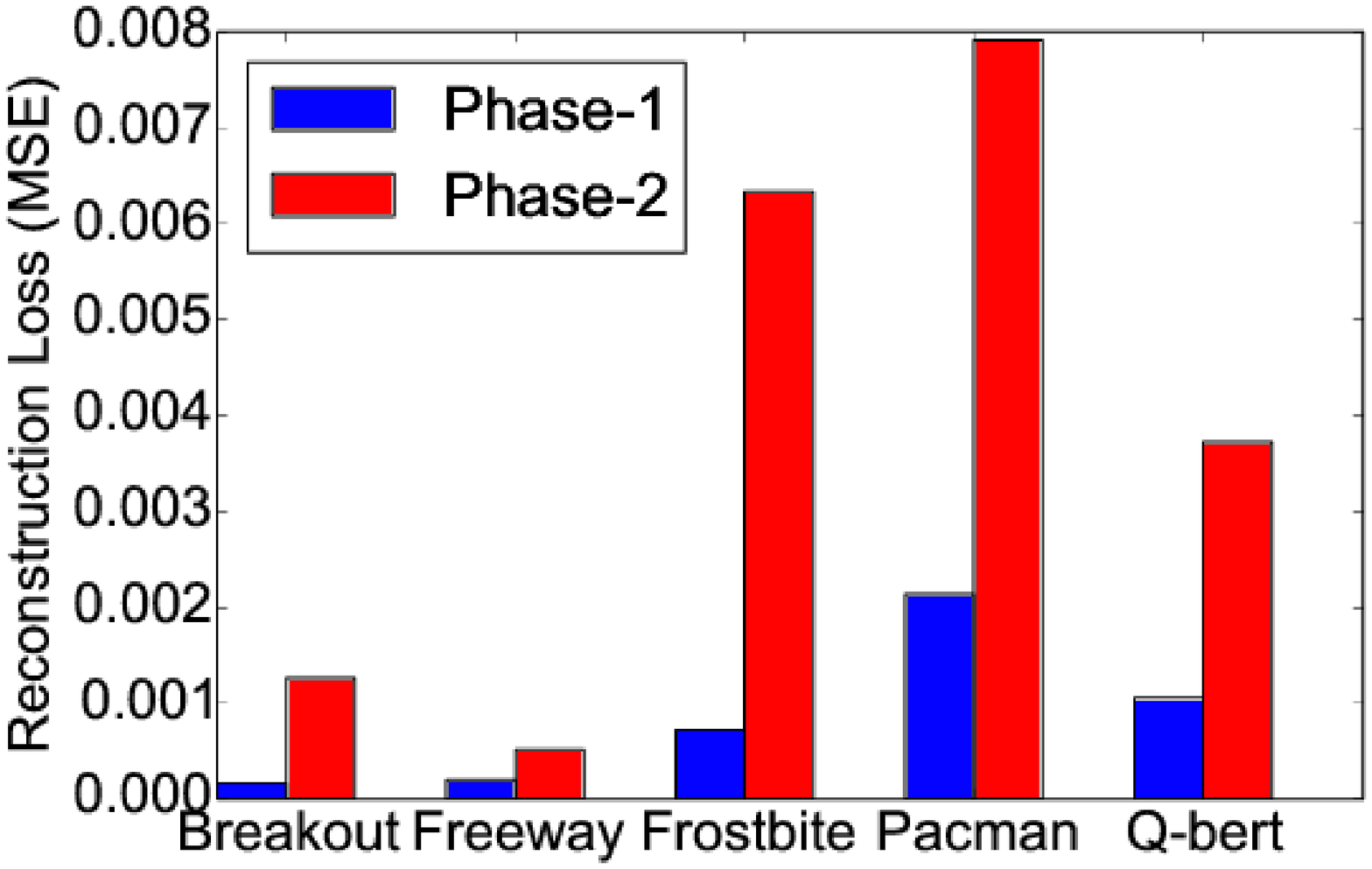}}  \\ 

\caption{Comparison of the code loss and the frame reconstruction loss (MSE) for autoencoder after the training of phase 1 \& phase 2.}
\vspace{-1.5mm}
\label{fig:hash:loss}
\end{figure}

\vspace{-3mm}
Overall, it is extremely challenging to match the state codes for the \textit{predicted} frames and their corresponding \textit{seen} frames while maintaining a satisfying reconstruction performance. We demonstrate this in Figure~\ref{fig:hash:loss} (a) by showing the \textit{code loss}, which is measured in terms of the number of mismatch in binary codes between a pair of predicted frame and its corresponding ground-truth frame. The presented result is derived by averaging over 10,000 pairs of codes. First, the result shows that without the second phase, it is impossible to perform hashing with autoencoder trained only by the reconstruction loss, since the average \textit{code losses} are above 1 in all the domains and with distinct hash codes, the count values returned from querying the hash table are meaningless. Second, the result shows that after the training of the second phase, the \textit{code loss} is significantly reduced. \linebreak

\begin{table*}[ht!]

  \caption{Performance score for the proposed approach and baseline RL approaches.}

  \label{table:RL_result}
  \label{RL_result}
  \begin{threeparttable}
  \centering

  \scalebox{0.88}{
  \makebox[\textwidth]{
  \begin{tabular}{llllll}
    \toprule
    Model     & \textbf{Breakout} & \textbf{Freeway}     &  \textbf{Frostbite}  & \textbf{Ms-Pacman} & \textbf{Q-bert} \\
    \midrule
    \textbf{DQN-Random}      & 401.2  & 30.9     &  328.3       &  2281       &    3876   \\
    \textbf{A3C} & 432.42 & 0 & 283.99 & 2327.8 & 19175.72 \\ \midrule
    \textbf{A3C-CTS} & \textbf{473.93} & 30.48 & 325.42 & 2401.04 & \textbf{19257.55} \\
    \textbf{pixelCNN} & 448.2 & 31.7 & 1480 & 2489.3 & 5876 \\

    \textbf{DQN-Informed}       & 0.93   & 32.2    &  1287.01      &  2522      &    8238     \\ \midrule
    \textbf{DQN-Informed-Hash}   & 451.93   &  \textbf{33.92  } &\textbf{1812.10  }  & \textbf{3526.60  }  &    8827.83  \\
    \bottomrule
  \end{tabular}
  }
  }
  \end{threeparttable}
\end{table*}

We also show the reconstruction errors measured in terms of MSE after the training of the two phases for each domain in Figure~\ref{fig:hash:loss} (b). By incorporating the code matching loss, the reconstruction behavior for the autoencoder receives slightly negative effect. A comparison of frame reconstruction effect after the training of the two phases are shown in Figure~\ref{fig:all_frames}.
It is shown that after training to match the state codes, the reconstructed frames are slightly blurred, but still able to reflect the essential features in each problem domain.

Moreover, we use \textit{Breakout} as an illustrative example to demonstrate the hashing can generate meaningful hash codes for \textit{predicted} future frames (see Figure ~\ref{fig:breakout:hash}). For a given ground-truth frame, we show the predicted frames with length 5 for taking each action. It can be found that different actions lead to different trajectories of board positions. For the hash codes, as three of the actions, \textit{no-op}, \textit{fire} and \textit{left}, lead to little change in frames, most of the future frames are hashed into the same code. The action \textit{right} leads to the most significant changes in board position, so the codes for future frames are much more distinct than the rest. Meanwhile, Figure~\ref{fig:breakout:hash} also shows that the multi-step prediction model generates realistic future frames.

\subsection{Evaluation on Informed Exploration}
We integrate the proposed informed exploration framework into DQN algorithm~\cite{nature} and compare with the following baselines: (1) DQN that performs $\epsilon$-greedy with uniform random action selection, denoted by DQN-Random; (2) A3C~\cite{mnih2016asynchronous}, (3)A3C with density model on Atari features~\cite{bellemare2016unifying}, (4) exploration with pixelCNN-based density model~\cite{ostrovski2017count}, (5) the state-of-the-art informed exploration approach proposed in~\cite{NIPS2015_5859}, denoted by {DQN-Informed}. Our proposed approach is denoted by {DQN-Informed-Hash}. We use $q = 3$ as the length for hashing over future frames. We report the result in Table~\ref{table:RL_result}.

Among all the test domains, {DQN-Informed-Hash} outperforms {DQN-Informed}, and there are significant performance gains observed in each domain. Note that in \textit{Breakout}, the agent fails to progress with {DQN-Informed} and always scores almost 0. It may due to that the kernel-based pixel distance evaluation metric used in {DQN-Informed} encourages the agent to explore states that is dissimilar from the recent history, which is insufficient to let the agent explore. Note that DQN-Informed-Hash demonstrates the superior performance with a \textit{deterministic} exploration mechanism. It indicates that counting over the predicted future frames could provide a meaningful direction for exploration.

\vspace{-0.5mm}
\section{Conclusion}

In this paper, we propose an informed exploration framework for deep RL domains with discrete action space. By incorporating a deep action-conditional prediction model over future transitions and a hashing mechanism based on a deep autoencoder model and LSH, we enable the agent to predict over the future trajectories and intuitively evaluate the novelty for each future action direction based on the hashing result. Empirical results on Atari 2600 domain show that the proposed informed exploration framework could significantly improve the exploration efficiency in several challenging deep RL domains.

\subsubsection*{Acknowledgments}
This work is funded by NTU Singapore Nanyang Assistant Professorship (NAP) grant M4081532.020, MOE AcRF Tier-1 grant 2016-T1-001-159, and Microsoft Research Asia.

\bibliographystyle{named}
\bibliography{ijcai17}

\begin{thebibliography}{}

\bibitem[\protect\citeauthoryear{Achiam and Sastry}{2017}]{joshua2017ICLR}
Joshua Achiam and Shankar Sastry.
\newblock Surprise-based intrinsic motivation for deep reinforcement learning.
\newblock {\em arXiv:1703.01732}, 2017.

\bibitem[\protect\citeauthoryear{{Bellemare} \bgroup \em et al.\egroup
  }{2013}]{bellemare13arcade}
M.~G. {Bellemare}, Y.~{Naddaf}, J.~{Veness}, and M.~{Bowling}.
\newblock The arcade learning environment: An evaluation platform for general
  agents.
\newblock {\em Journal of Artificial Intelligence Research}, 47:253--279, Jun
  2013.

\bibitem[\protect\citeauthoryear{Bellemare \bgroup \em et al.\egroup
  }{2016}]{bellemare2016unifying}
Marc Bellemare, Sriram Srinivasan, Georg Ostrovski, Tom Schaul, David Saxton,
  and Remi Munos.
\newblock Unifying count-based exploration and intrinsic motivation.
\newblock In {\em NIPS}, pages 1471--1479, 2016.

\bibitem[\protect\citeauthoryear{Blundell \bgroup \em et al.\egroup
  }{2017}]{blundell2016model}
Charles Blundell, Benigno Uria, Alexander Pritzel, Yazhe Li, Avraham Ruderman,
  Joel~Z Leibo, Jack Rae, Daan Wierstra, and Demis Hassabis.
\newblock Model-free episodic control.
\newblock In {\em ICML}, 2017.

\bibitem[\protect\citeauthoryear{Charikar}{2002}]{simHash}
Moses~S Charikar.
\newblock Similarity estimation techniques from rounding algorithms.
\newblock In {\em STOC}, pages 380--388. ACM, 2002.

\bibitem[\protect\citeauthoryear{Houthooft \bgroup \em et al.\egroup
  }{2016}]{houthooft2016vime}
Rein Houthooft, Xi~Chen, Yan Duan, John Schulman, Filip De~Turck, and Pieter
  Abbeel.
\newblock Vime: Variational information maximizing exploration.
\newblock In {\em NIPS}, pages 1109--1117, 2016.

\bibitem[\protect\citeauthoryear{Kingma and Ba}{2014}]{adam}
Diederik~P. Kingma and Jimmy Ba.
\newblock Adam: A method for stochastic optimization.
\newblock {\em CoRR}, abs/1412.6980, 2014.

\bibitem[\protect\citeauthoryear{Kingma and Welling}{2014}]{kingma2013auto}
Diederik~P Kingma and Max Welling.
\newblock Auto-encoding variational bayes.
\newblock In {\em ICLR}, 2014.

\bibitem[\protect\citeauthoryear{Lillicrap \bgroup \em et al.\egroup
  }{2016}]{lillicrap2015continuous}
Timothy~P Lillicrap, Jonathan~J Hunt, Alexander Pritzel, Nicolas Heess, Tom
  Erez, Yuval Tassa, David Silver, and Daan Wierstra.
\newblock Continuous control with deep reinforcement learning.
\newblock In {\em ICLR}, 2016.

\bibitem[\protect\citeauthoryear{Martin \bgroup \em et al.\egroup
  }{2017}]{martin2017count}
Jarryd Martin, Suraj~Narayanan Sasikumar, Tom Everitt, and Marcus Hutter.
\newblock Count-based exploration in feature space for reinforcement learning.
\newblock In {\em IJCAI}, 2017.

\bibitem[\protect\citeauthoryear{Mnih \bgroup \em et al.\egroup
  }{2015}]{nature}
V.~Mnih, K.~Kavukcuoglu, D.~Silver, A.~a~Rusu, J.~Veness, M.~G. Bellemare,
  A.~Graves, M.~Riedmiller, A.~K. Fidjeland, G.~Ostrovski, S.~Petersen,
  A.~Sadik C.~Beattie, I.~Antonoglou, D.~Kumaran H.~King, D.~Wierstra, S.~Legg,
  and D.~Hassabis.
\newblock Human-level control through deep reinforcement learning.
\newblock {\em Nature}, 2015.

\bibitem[\protect\citeauthoryear{Mnih \bgroup \em et al.\egroup
  }{2016}]{mnih2016asynchronous}
Volodymyr Mnih, Adria~Puigdomenech Badia, Mehdi Mirza, Alex Graves, Timothy~P
  Lillicrap, Tim Harley, David Silver, and Koray Kavukcuoglu.
\newblock Asynchronous methods for deep reinforcement learning.
\newblock In {\em ICML}, 2016.

\bibitem[\protect\citeauthoryear{Oh \bgroup \em et al.\egroup
  }{2015}]{NIPS2015_5859}
Junhyuk Oh, Xiaoxiao Guo, Honglak Lee, Richard~L Lewis, and Satinder Singh.
\newblock Action-conditional video prediction using deep networks in atari
  games.
\newblock In {\em NIPS}, pages 2845--2853. 2015.

\bibitem[\protect\citeauthoryear{Osband \bgroup \em et al.\egroup
  }{2016}]{osband2016deep}
Ian Osband, Charles Blundell, Alexander Pritzel, and Benjamin Van~Roy.
\newblock Deep exploration via bootstrapped dqn.
\newblock In {\em NIPS}, pages 4026--4034, 2016.

\bibitem[\protect\citeauthoryear{Ostrovski \bgroup \em et al.\egroup
  }{2017}]{ostrovski2017count}
Georg Ostrovski, Marc~G Bellemare, Aaron van~den Oord, and R{\'e}mi Munos.
\newblock Count-based exploration with neural density models.
\newblock In {\em ICML}, 2017.

\bibitem[\protect\citeauthoryear{Parisotto \bgroup \em et al.\egroup
  }{2016}]{Actor}
Emilio Parisotto, Jimmy Ba, and Ruslan Salakhutdinov.
\newblock Actor-mimic deep multitask and transfer reinforcement learning.
\newblock In {\em ICLR}, 2016.

\bibitem[\protect\citeauthoryear{Pathak \bgroup \em et al.\egroup
  }{2017}]{pathak2017curiosity}
Deepak Pathak, Pulkit Agrawal, Alexei~A Efros, and Trevor Darrell.
\newblock Curiosity-driven exploration by self-supervised prediction.
\newblock In {\em ICML}, 2017.

\bibitem[\protect\citeauthoryear{Rusu \bgroup \em et al.\egroup
  }{2016}]{Distillation}
Andrei~A. Rusu, Sergio~Gomez Colmenarejo, Caglar Gulcehre, Guillaume
  Desjardins, James Kirkpatrick, Razvan Pascanu, Volodymyr Mnih, Koray
  Kavukcuoglu, and Raia Hadsell.
\newblock Policy distillation.
\newblock In {\em ICLR}, 2016.

\bibitem[\protect\citeauthoryear{Schaul \bgroup \em et al.\egroup
  }{2016}]{Priority}
Tom Schaul, John Quan, Ioannis Antonoglou, and David Silver.
\newblock Prioritized experience replay.
\newblock In {\em ICLR}, 2016.

\bibitem[\protect\citeauthoryear{Stadie \bgroup \em et al.\egroup
  }{2015}]{incentivize}
Bradly~C Stadie, Sergey Levine, and Pieter Abbeel.
\newblock Incentivizing exploration in reinforcement learning with deep
  predictive models.
\newblock {\em arXiv:1507.00814}, 2015.

\bibitem[\protect\citeauthoryear{Strehl and Littman}{2008}]{strehl2008analysis}
Alexander~L Strehl and Michael~L Littman.
\newblock An analysis of model-based interval estimation for markov decision
  processes.
\newblock {\em Journal of Computer and System Sciences}, 74(8):1309--1331,
  2008.

\bibitem[\protect\citeauthoryear{Sutton and
  Barto}{1998}]{sutton1998reinforcement}
Richard~S Sutton and Andrew~G Barto.
\newblock {\em Reinforcement learning: An introduction}, volume~1.
\newblock MIT press Cambridge, 1998.

\bibitem[\protect\citeauthoryear{Tang \bgroup \em et al.\egroup }{2017}]{Tang}
Haoran Tang, Rein Houthooft, Davis Foote, Adam Stooke, Xi~Chen, Yan Duan, John
  Schulman, Filip De~Turck, and Pieter Abbeel.
\newblock \# exploration: A study of count-based exploration for deep
  reinforcement learning.
\newblock In {\em NIPS}, 2017.

\bibitem[\protect\citeauthoryear{Thrun}{1992}]{thrun1992efficient}
Sebastian~B Thrun.
\newblock Efficient exploration in reinforcement learning.
\newblock 1992.

\bibitem[\protect\citeauthoryear{Wang \bgroup \em et al.\egroup
  }{2016}]{Dueling}
Ziyu Wang, Tom Schaul, Matteo Hessel, Hado van Hasselt, Marc Lanctot, and Nando
  de~Freitas.
\newblock Dueling network architectures for deep reinforcement learning.
\newblock In {\em ICML}, pages 1995--2003, 2016.

\bibitem[\protect\citeauthoryear{Yin and Pan}{2017}]{yin2017knowledge}
Haiyan Yin and Sinno~Jialin Pan.
\newblock Knowledge transfer for deep reinforcement learning with hierarchical
  experience replay.
\newblock In {\em AAAI}, pages 1640--1646, 2017.

\end{thebibliography}

\end{document}